\documentclass[10pt,twocolumn,letterpaper, table]{article}

\usepackage{cvpr}              
\usepackage[T1]{fontenc}
\usepackage[utf8]{inputenc}
\usepackage{fontawesome5}
\usepackage[dvipsnames]{xcolor}

\definecolor{cvprblue}{rgb}{0.21,0.49,0.74}
\usepackage[pagebackref,breaklinks,colorlinks,citecolor=cvprblue]{hyperref}

\def\paperID{9020} 
\def\confName{CVPR}
\def\confYear{2024}

    \title{CoGS \faIcon{cogs}: Controllable Gaussian Splatting}

\author{Heng Yu$^{1}$ \quad Joel Julin$^{1}$  \quad Zolt\'{a}n \'{A}. Milacski$^{1}$\quad Koichiro Niinuma$^{2}$ \quad L\'{a}szl\'{o} A. Jeni$^{1}$ \vspace{4pt}\\
	$^1$Robotics Institute, Carnegie Mellon University \quad
    $^2$Fujitsu Research of America \\
    {\tt\small \{hengyu, jjulin, zmilacsk\}@andrew.cmu.edu} \quad {\tt\small kniinuma@fujitsu.com} \quad {\tt\small laszlojeni@cmu.edu} \\
}

\begin{document}
\twocolumn[{%
\renewcommand\twocolumn[1][]{#1}%
\maketitle
\vspace{-1.2cm}
\begin{center}
\url{https://cogs2024.github.io}
\end{center}
\vspace{0.1cm}
\begin{center}
    \centering
    \captionsetup{type=figure}
    \begin{minipage}{0.33\textwidth}
    \centering
    \vspace{-0.5cm}
    \includegraphics[width=1\textwidth]{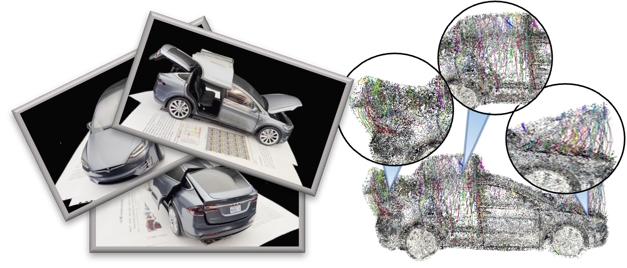}\\
    (a) Dynamic 3D Gaussians \\
    \end{minipage}
    \begin{minipage}{0.33\textwidth}
    \centering
    \vspace{-0.1cm}
    \includegraphics[width=0.85\textwidth]{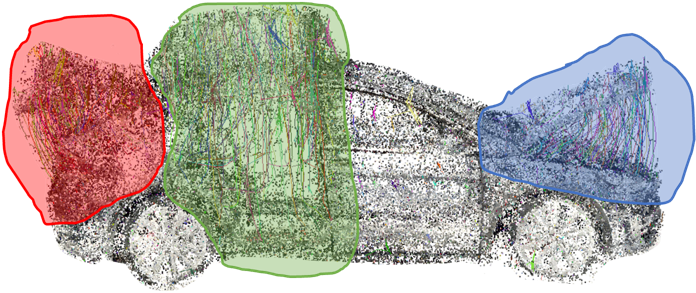} \\
    (b) 2D-to-3D Mask Projection  \\
    \end{minipage}
    \hspace{-1cm}
    \begin{minipage}{0.33\textwidth}
    \centering
    \vspace{-0.46cm}
    \includegraphics[width=0.85\textwidth]{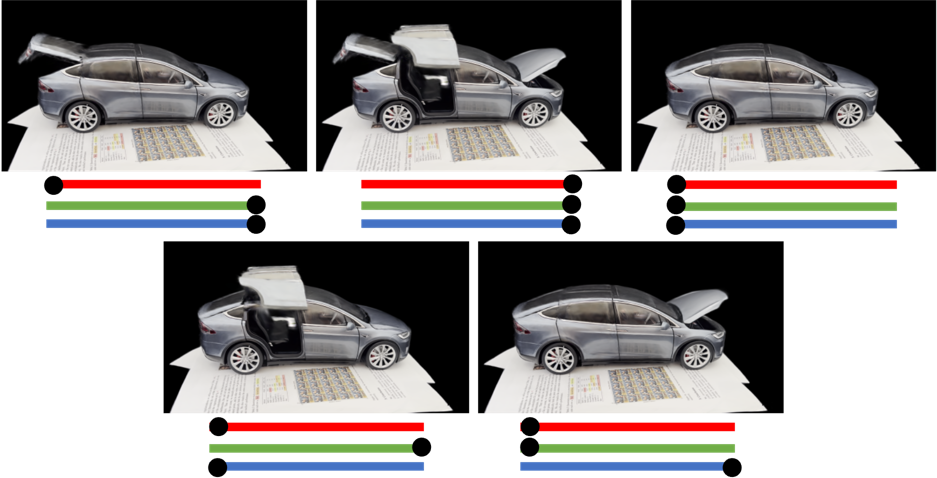} \\
    (c) Attribute Control \\
    \end{minipage}
    \captionof{figure}{From a set of monocular images capturing a moving scene, a dynamic 3D representation is learned using time-varying Gaussians (a). Then the articulated parts (depicted with the trajectories of the motion) are identified using masking (b). This allows for learning a fine-scale, per-Gaussian level of control (c). The approach is capable of synthesizing novel configurations not present in the original sequence, for example, independently opening the hood, trunk, and doors of the toy car. 
    }
    \label{fig:fancy}
\end{center}%
}]
\maketitle
\begin{abstract}
Capturing and re-animating the 3D structure of articulated objects present significant barriers. On one hand, methods requiring extensively calibrated multi-view setups are prohibitively complex and resource-intensive, limiting their practical applicability. On the other hand, while single-camera Neural Radiance Fields (NeRFs) offer a more streamlined approach, they have excessive training and rendering costs. 3D Gaussian Splatting would be a suitable alternative but for two reasons. Firstly, existing methods for 3D dynamic Gaussians require synchronized multi-view cameras, and secondly, the lack of controllability in dynamic scenarios. We present CoGS, a method for Controllable Gaussian Splatting, that enables the direct manipulation of scene elements, offering real-time control of dynamic scenes without the prerequisite of pre-computing control signals. We evaluated CoGS using both synthetic and real-world datasets that include dynamic objects that differ in degree of difficulty.
In our evaluations, CoGS consistently outperformed existing dynamic and controllable neural representations in terms of visual fidelity.
\end{abstract}
\section{Introduction}
\label{sec:intro}

Recent advancements in machine vision have significantly enhanced our ability to interpret and reconstruct 3D structures from 2D observations. This progress is largely due to the development of coordinate networks, such as Neural Radiance Fields (NeRF) \cite{mildenhall2021nerf} and its variants \cite{barron2021mip,mildenhall2022nerf,wang2021neus,martin2021nerf}, which have revolutionized high-fidelity novel-view synthesis and scene representation. NeRFs, however, primarily focus on static scenes and their implicit representation poses challenges in direct scene manipulation. Addressing dynamic scenes involves additional complexities, as seen in various extensions \cite{pumarola2021d,tretschk2021non,gafni2021dynamic,park2021nerfies}, which often require intricate mechanisms to adapt to scene deformations.

In contrast to the implicit nature of NeRFs, our work centers on Gaussian Splatting (GS), a method characterized by its explicit representation. Building on this concept of 3D GS, which employs 3D Gaussians for scene modeling, we extend this approach to dynamic and controllable scenarios. The explicit nature of GS \cite{kerbl20233d} not only facilitates more efficient rendering compared to the computationally intensive ray-casting and numerical integration of NeRFs but also significantly simplifies the manipulation of scene elements, offering direct control over the Gaussians.

We propose a novel framework that adapts GS for dynamic environments captured by a monocular camera, integrating control mechanisms that allow for intuitive and straightforward manipulation of scene elements. This development addresses the limitations of NeRFs in terms of computational complexity and challenges in scene manipulation due to their implicit representation. By leveraging the explicit 3D Gaussian representations and combining them with advanced control techniques, our method opens new avenues for real-time, high-fidelity scene rendering and manipulation, particularly relevant in fields such as virtual reality, augmented reality, and interactive media.

\section{Related Works}
\label{sec:formatting}


\subsection{Dynamic NeRFs}


NeRFs have shown remarkable capabilities in synthesizing novel views of static scenes. The extension of these techniques to dynamic deformable domains has been a focal point of recent research \cite{pumarola2021d, tretschk2021non, gafni2021dynamic, park2021nerfies, park2021hypernerf}. A critical aspect in these advancements is the effective modeling of deformation. Approaches vary: some employ translational deformation fields with temporal positional encoding, as seen in D-NeRF \cite{pumarola2021d} and NR-NeRF \cite{tretschk2021non}, while others utilize rigid body motion fields, exemplified by Nerfies \cite{park2021nerfies} and HyperNeRF \cite{park2021hypernerf}. Notably, HyperNeRF \cite{park2021hypernerf} introduces a hyperspace representation to capture topological variations. Additionally, optical flow has been explored as a method for deformation regularization \cite{du2021neural, gao2021dynamic, li2021neural, wang2023flow}. Research on dynamic scenes often also addresses the use of multiple synchronized cameras \cite{li2022neural, wang2022fourier, li2022streaming, wang2023mixed} and focuses on accelerating both training \cite{deng2022depth, sun2022direct, muller2022instant, fridovich2022plenoxels, chen2022tensorf, yariv2023bakedsdf} and inference processes \cite{yu2023dylin, mubarik2023hardware, cao2023real, lin2022efficient}.


\subsection{Controllable NeRFs}


In addition to dynamic NeRFs, another area of research is re-animation of dynamic scenes \cite{kania2022conerf, garbin2022voltemorph, lazova2023control, athar2022rignerf}. CoNeRF \cite{kania2022conerf}, which is closely related to our work, introduces manually labeled control signals and control area masks into the hyperspace framework proposed by HyperNeRF \cite{park2021hypernerf}. Building on this, CoNFies \cite{yu2023confies} advances the concept to a fully automatic system, also achieving accelerated rendering speeds by distilling knowledge to a student Light Field Network (LFN) \cite{yu2023dylin}. However, a limitation of these approaches is the necessity for pre-computed or labeled control signals and masks. This requirement, stemming from the implicit representation of hyperspace or neural radiance fields, significantly restricts their applicability in broader contexts.

\subsection{Gaussian Splatting}


Recently, 3D GS has emerged as a promising technique \cite{kerbl20233d}. This method explicitly models scenes using 3D Gaussians, characterized by parameters such as mean, variance, color, and density. Unlike NeRF's ray-based rendering, it employs rasterization, leading to faster training and rendering while enhancing image quality. Initially focused on static scenes, 3D GS has been extended to dynamic scenarios by concurrent research \cite{wu20234d, yang2023deformable, yang2023real}, aligning with our work's core ideas. These extensions often adopt additional networks to model dynamic behavior, reminiscent of approaches in dynamic NeRFs. However, they do not fully leverage the explicit nature of the Gaussians. Our work distinguishes itself by introducing controllability into dynamic GS, taking full advantage of the explicit Gaussian representations for manipulation.

\begin{figure*}[t]
    \centering
    \includegraphics[width=0.7\linewidth]{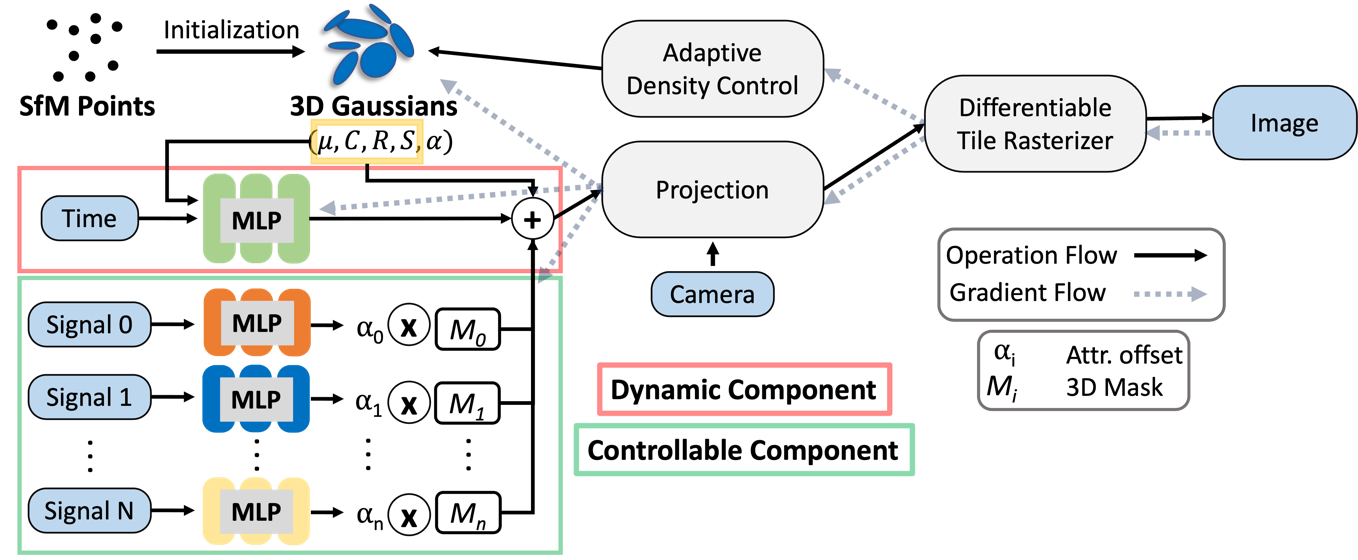}
    \caption{CoGS Overview. CoGS consists of two parts: Dynamic GS and Controllable GS. For Dynamic GS, an offset is learned for $(\mu, C, R, S)$ by separate MLPs (only one shown in figure). To extend to controllable scenarios, signals extracted from the dynamic model are used to obtain attribute offsets, which are then masked to affect the desired control region.}
    \label{fig:pipeline}
\end{figure*}
\section{Methods}
In order to realize controllable GS, it is essential to first establish a GS framework capable of modeling dynamic scenes. This chapter is dedicated to unfolding this process in two distinct phases: initially, we introduce the concept and methodology of dynamic GS.
Subsequently, we build upon this foundation to evolve these methods into a controllable framework, thereby enhancing their adaptability and applicability in dynamic scene modeling.
\subsection{Dynamic Gaussian Splatting}
To represent dynamic scenes and ultimately introduce fine-scale attribute control with 3D Gaussians, we utilize the differentiable Gaussian rasterization pipeline proposed by \cite{kerbl20233d}. Specifically, we directly follow the method therein and then augment its static properties with deformation fields to model scene dynamics.

\subsubsection{Differentiable Rasterization of 3D Gaussians}
\label{sec:diff_rast}
Each 3D Gaussian is defined by a full 3D covariance matrix $\mathbf{\Sigma}$, position (mean) $\mathbf{\mu}$, opacity $\alpha$, and color represented via spherical harmonics (SH). To render these 3D Gaussians, projecting them from 3D to 2D Gaussians, we follow the procedure outlined in \cite{zwicker-a} to obtain the view space covariance matrix $\mathbf{\Sigma'}$:
\begin{equation}
    \mathbf{\Sigma'} = \mathbf{J}\mathbf{W}\mathbf{\Sigma}\mathbf{W}^{T}\mathbf{J}^{T},
\end{equation}
where $\mathbf{W}$ is the view transform and $\mathbf{J}$ is the Jacobian of the affine approximation of the projective transformation.

Since the physical meaning of a covariance matrix is only valid if it is positive semi-definite, it  cannot be easily optimized to best represent a scene's radiance field \cite{kerbl20233d}. However, we can obscure this complexity by employing a parameterization that inherently maintains the positive semi-definiteness of the matrix. The covariance matrix $\Sigma$ can be decomposed into intuitive and optimizable components that correspond to an ellipsoid's scaling and orientation with rotation matrix $\mathbf{R}$ and scaling matrix $\mathbf{S}$:
\begin{equation}
    \mathbf{\Sigma} = \mathbf{R}\mathbf{S}\mathbf{S}^{T}\mathbf{R}^{T}.
\end{equation}
and optimize $\mathbf{R}$, $\mathbf{S}$ instead of $\mathbf{\Sigma}$.
 After projection, the Gaussians are sorted from front-to-back where the color $C$ is given by NeRF-like volumetric rendering along a ray:
 \begin{equation}
     \mathbf{C} = \sum^{N}_{i=1}T_{i}(1-exp(-\sigma_{i}\delta_{i}))\mathbf{c}_{i},
    \label{eq:render}
 \end{equation}
with:
\begin{equation}
    T_{i} = exp(-\sum_{j=1}^{i-1}\sigma_{j}\delta_j).
\end{equation}

During optimization, we adaptively control the density of the 3D Gaussians to best represent the scene. Throughout this process, the total number of Gaussians will change.
For a more comprehensive outline of this procedure, we kindly ask the readers to refer to \cite{kerbl20233d}.
\subsubsection{Optimization for Dynamic Scene Representation}
The defining parameters of each 3D Gaussian presuppose a static scene. Our approach bridges this gap to dynamic scenarios by learning independent deformation networks for each parameter. Additionally, we introduce multiple losses to maintain geometric consistency across time. The pipeline overview is presented in Fig. \ref{fig:pipeline}, with the dynamic component highlighted in red.

We initialize a set of 3D Gaussians from a Structure from Motion (SfM)~\cite{schonberger2016structure} point-cloud (or randomly selecting $N$ points within the scene box), each defined by the same parameters as in \ref{sec:diff_rast}. For the first 3000 iterations, our focus is exclusively on learning the static elements within the scene. This deliberate emphasis on stabilizing the static portions proves to be crucial for achieving high performance on the dynamic reconstruction. Establishing a robust static foundation lays the groundwork for a more accurate reconstruction of dynamic elements. During this phase, the deformation network (green MLP) does not update any parameters. Instead, these 3D Gaussians adhere to the same differentiable rasterization pipeline as covered in Section \ref{sec:diff_rast}.

In the subsequent phases, the deformation network is employed to update each parameter, tailoring them to the dynamic scene. Although not explicitly depicted in Fig. \ref{fig:pipeline}, we learn $j$ separate networks, one for each parameter. For  ($\mathbf{\mu_{i}}$, $\mathbf{C}_{i}$, $\mathbf{R_{i}}$, $\mathbf{S_{i}}$), we have a network $N_{j}$ such that:
\begin{equation}
    N_{j}(\mathbf{\mu_{i}}, t) = (\Delta \mathbf{ \mu_{i}}, \Delta \mathbf{C}_{i}, \Delta \mathbf{ R_{i}}, \Delta \mathbf{ S_{i}}),
\end{equation}
where $t$ is the current time step. Different from \cite{yang2023deformable}, we also learn an offset for color to account for any changes that may occur over time (e.g. shadows and reflections). The outputs from these networks are then added to the corresponding parameters, and the differentiable rasterization pipeline proceeds.

Learning offsets alone results in a method that is unaware of consistent trajectories and accurate movement. Thus, we employ our multiple regularization losses to further constrain this difficult problem. 

\begin{figure}[ht]
\centering
\begin{subfigure}{0.2\textwidth}
  \centering
  \includegraphics[width=0.77\linewidth]{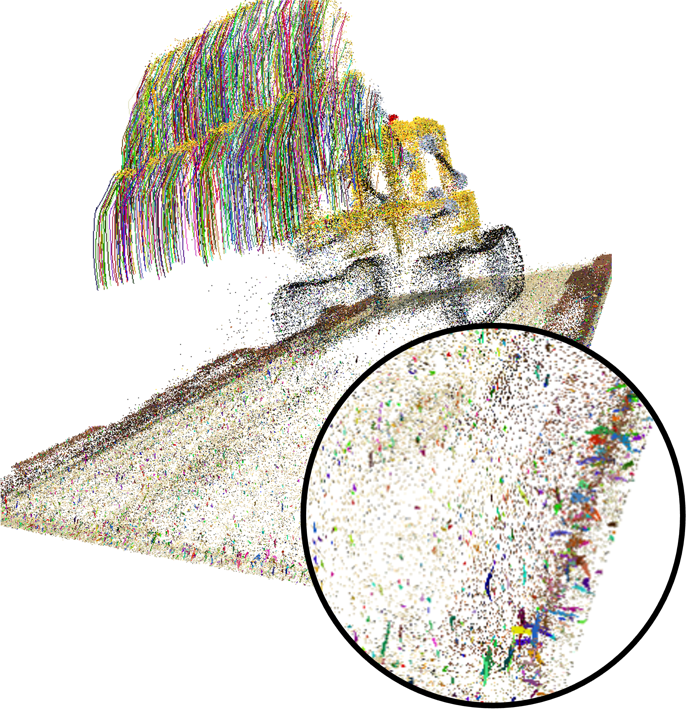}
  \caption{}
  \label{fig:with-traj}
\end{subfigure}%
\begin{subfigure}{0.2\textwidth}
  \centering
  \includegraphics[width=0.77\linewidth]{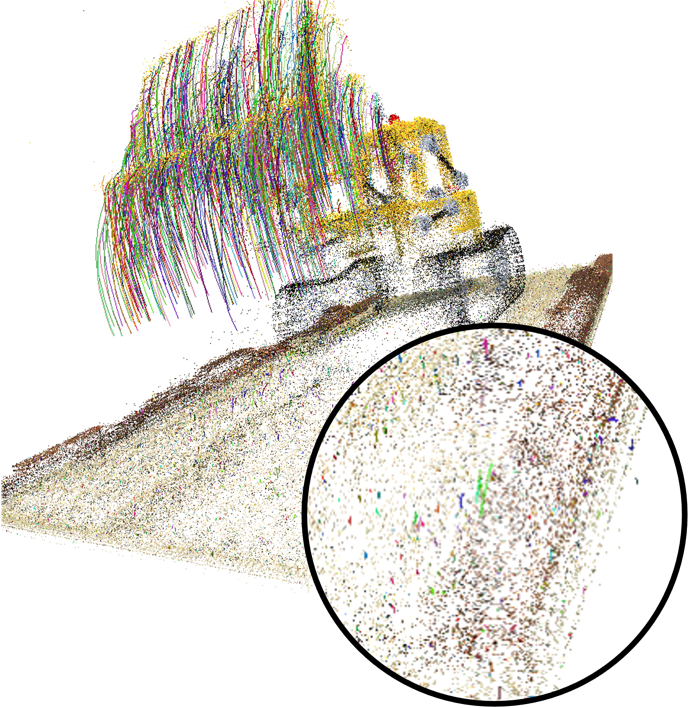}
  \caption{}
  \label{fig:no-traj}
\end{subfigure}
\caption{Lego synthetic scene visualized as a pointcloud of colored Gaussian centers. The smaller and fewer colored lines indicate less change in position over time. Adding $\mathcal{L^{\text{norm}}}$ stabilizes the static Gaussian's positions. (a) Without $\mathcal{L^{\text{norm}}}$. (b) With $\mathcal{L^{\text{norm}}}$.}
\label{fig:ablation_lego}
\end{figure}

For each time step, the mean of the normalized predicted position offsets ($\Delta \mathbf{\mu}$) is computed to ensure their consistency with one another. Specifically, we use this to localize position offsets.
\begin{equation}
    \mathcal{L}^{\text{norm}} = \frac{1}{N}\sum^{N}_{i=1}\|\Delta \mathbf{ \mu_{i}}\|.
\end{equation}
As shown in Fig. \ref{fig:ablation_lego}, the trajectories of static portions of the scene tend to stabilize with the addition of this loss.

After 15000 iterations, we enforce the remainder of our losses. Specifically, a local difference loss, denoted as $\mathcal{L}^{\text{diff}}$, is used to ensure the movement, or trajectory, for each Gaussian is consistent with its neighbors over time. This loss is formulated as follows:
\begin{equation}
    \mathcal{L}^{\text{diff}}_{i, j} = \left|\|\mathbf{\mu}_{i, j, t} - \mathbf{\mu_{i, t}}\| - \|\mathbf{\mu_{i, j, t-1}} - \mathbf{\mu_{i, t-1}}\|\right|.
\end{equation}
Here, $\mu_{i, j, t}$ represents the position of a nearest neighbor Gaussian $j$ to Gaussian $i$ at time $t$. Similarly, $\mu_{i, t}$ is the position of Gaussian $i$ at time $t$, and analogous notation is used for the time step $t-1$.

The overall local difference loss is then defined as the average over all Gaussians $i$ and their $k$-nearest neighbours:
\begin{equation}
    \mathcal{L}^{\text{diff}} = \frac{1}{k|G|}\sum_{i \in G} \sum_{j \in \text{knn}_{i;k}}\mathcal{L}^{\text{diff}}_{i, j}.
\end{equation}
Demonstrated in Fig. \ref{fig:with-traj}, this loss yields a much more consistent dynamic representation when compared to without it as shown in \ref{fig:no-traj}. 
\begin{figure}[ht]
\centering
\begin{subfigure}{0.22\textwidth}
  \centering
  \includegraphics[width=0.77\linewidth]{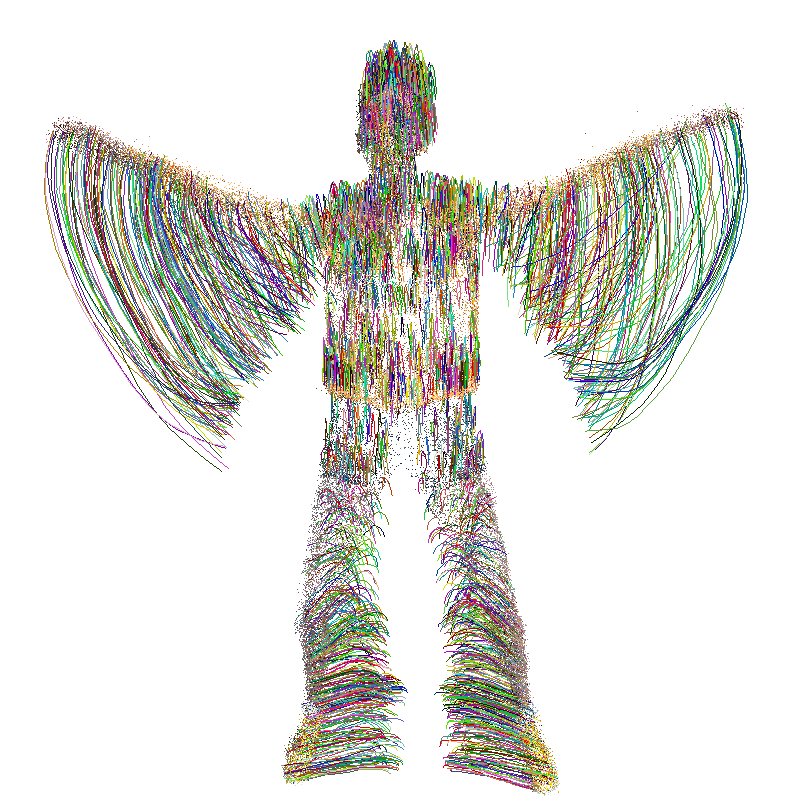}
  \caption{}
  \label{fig:with-traj}
\end{subfigure}%
\begin{subfigure}{0.22\textwidth}
  \centering
  \includegraphics[width=0.77\linewidth]{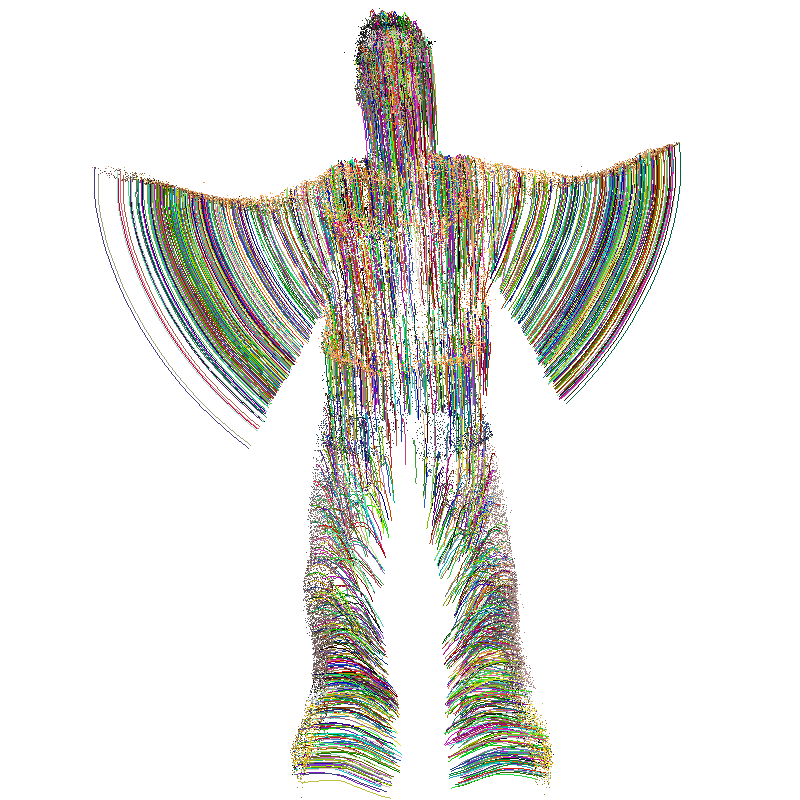}
  \caption{}
  \label{fig:no-traj}
\end{subfigure}
\caption{Jumping Jack synthetic scene visualized as a pointcloud of colored Gaussian centers. The smaller and fewer colored lines indicate less change in position over time. Adding $\mathcal{L^{\text{diff}}}$ stabilizes the 3D Gaussian's trajectories. (a) Without $\mathcal{L^{\text{diff}}}$. (b) With $\mathcal{L^{\text{diff}}}$.}
\label{fig:ablation}
\end{figure}

The next two loss functions are directly taken from \cite{luiten2023dynamic}, for more in depth details, please refer to their paper. Each of these losses are assigned a weight through an unnormalized Gaussian weighting factor:
\begin{equation}
    w_{i, j} = \exp{(-\lambda_{w}\| \mu_{j, 0} - \mu_{i, 0}\|^{2}_{2})}
\end{equation}
The distance between each Gaussian's position is computed at the first time step, and then it is fixed for the remaining part of the sequence. In doing this, each of the following losses are explicitly locally enforced.

Using this weighting scheme, a local-rigidity loss is employed, denoted as $\mathcal{L}^{\text{rigid}}$, defined as follows:
\begin{equation}
    \mathcal{L}^{\text{rigid}}_{i, j} = w_{i, j}||(\mu_{j, t-1} - \mu_{i, t-1}) - \mathbf{R}_{i, t-1}\mathbf{R}^{-1}_{i, t}(\mu_{j, t} - \mu_{i, t}||_{2},
\end{equation}
\begin{equation}
    \mathcal{L}^{\text{rigid}} = \frac{1}{k|G|}\sum_{i \in G} \sum_{j \in \text{knn}_{i;k}}\mathcal{L}^{\text{rigid}}_{i, j}.
\end{equation}
This loss ensures that for each Gaussian $i$, neighboring Gaussians $j$ should move in a manner consistent with the rigid body transform of the coordinate system over time. 

Additionally, we incorporate a rotational loss $\mathcal{L}^{\text{rot}}$ to maintain consistency in rotations among nearby Gaussians across different time steps. This is expressed as:
\begin{equation}
    \mathcal{L}^{\text{rot}} = \frac{1}{k|G|}\sum_{i \in G}\sum_{j\in \text{knn}_{i;k}} w_{i, j}\|\hat{q}_{j, t} \hat{q}_{j, t-1}^{-1} - \hat{q}_{i, t}\hat{q}_{i, t-1}^{-1} \|_{2},
\end{equation}
where $\hat{q}$ is the normalized quaternion rotation of each Gaussian. The same $k$-nearest neighbors are used, as in the preceding losses. 

Each of the described losses is critical to success at dynamic scene reconstruction, as there exist multiple facets that require precise constraints.

\subsection{Controllable Gaussian Splatting}
Having established the framework for dynamic GS, we can now extend it to accommodate controllable scenarios as shown in Fig. \ref{fig:pipeline}. This extension is facilitated by its explicit, Gaussian-based representation. The comprehensive pipeline of our approach comprises four key steps:

\begin{enumerate}
    \item \textbf{Building a Dynamic GS Model:} As previously discussed, this foundational step establishes the groundwork for subsequent extensions.
    \item \textbf{3D Mask Generation:} This step involves translating two-dimensional mask data into a three-dimensional context, bridging the gap between simple representations and complex spatial models. The 2D mask is either annotated manually quite easily or automatically inferred by existing methods.
    \item \textbf{Control Signal Extraction:} A pivotal phase where control signals are identified and extracted manually or automatically from explict Gaussian sets, serving as the primary drivers for scene manipulation.
    \item \textbf{Control Signal Re-Alignment:} The final phase, which entails adjusting and aligning the control signals to ensure their seamless integration and responsiveness within the dynamic model.
\end{enumerate}

In the following sections, we will explore the details of the last three steps, elucidating their roles in enhancing the overall efficacy and controllability of our dynamic GS. 

\subsubsection{3D Mask Generation}

To delineate the controllable set of Gaussians, we introduce a mask vector \( m_i \in \mathbb{R}^L \) for each Gaussian, where \( L \) denotes the number of attributes to be controlled. The straightforward approach of selecting these in 3D introduces two major challenges: the complexity of manually labeling Gaussian positions for each attribute and the difficulty in achieving an exact fine-grained boundary for the 3D point set, potentially leading to control artifacts.

Addressing these challenges, we propose an effective method to obtain the mask vector \( m \). We start by acquiring \( K \) 2D masks for the 2D frames, where \( K \) can vary from all frames (for scenarios with available automatic mask generation methods like face recognition~\cite{yu2023dylin}, as illustrated in Fig.~\ref{fig:face_mask}) to a single frame (which can be manually labeled, as demonstrated in Figs.~\ref{fig:ball_mask} \& \ref{fig:torchchoco_mask} \& \ref{fig:car_mask}). After the 2D mask acquisition, we perform a 2D-to-3D mask projection. A practical method involves associating the 3D point with the corresponding 2D pixel, utilizing depth maps and camera poses as in~\cite{luiten2023dynamic}. However, this method falls short of attaining the fine-grained boundary of the controllable part due to the splatting process \ref{sec:diff_rast}.

Therefore, we suggest a learning process to obtain the mask vector for each Gaussian. We allocate a learnable mask tensor \( m_i \in \mathbb{R}^L \) to each Gaussian and implement a softmax operation to normalize the sum of the tensor to 1, aiming for a categorical distribution. For rendering the 2D mask $M$ from the 3D $m_i$, we use the same GS equation, referred to as Eq.~\ref{eq:render}, employing the same point \( \mu \), rotation matrix \( R \), and scaling matrix \( S \), 
except that we set the color for each Gaussian as a constant (1), and take the mask tensor as opacity. This rendered 2D mask is supervised using the ground-truth mask. Importantly, rather than enforcing an exact match between the rendered and ground-truth masks for each control area, we focus on ensuring that the rendered mask has no impact (is black) on other control areas as in Eq.~\ref{eq:Lmask}, significantly reducing artifacts at the boundaries. $M_i$ is the rendering 2D mask for the $i$th attribute and $M^{gt}_i$ is the corresponding ground truth. Here we want the rendering mask $M_i$ to ideally be black on other controllable areas so as to make no effect on these parts.
\begin{equation}
  \mathcal{L}^{\text{mask}} = \sum^{L}_{i=1}\|(1-M_i)-\sum^{L}_{j=1, j\neq i}M^{gt}_j\| \cdot \sum^{L}_{j=1, j\neq i}M^{gt}_j.
  \label{eq:Lmask}
\end{equation}

In this step, we maintain all other learnable weights and tensors, except for the mask tensor, as fixed.

\subsubsection{Control Signal Extraction}


Our method uniquely eliminates the need for pre-computed control signals, significantly expanding its range of applications. This is accomplished by unsupervised learning of the control signal directly from the Gaussians. The first step involves selecting a set of Gaussians, denoted as \( G \), which represents movement within the control part, as indicated by the previously learned mask \( m \). This set \( G \) can be either manually selected in 3D or automatically based on movement trajectories, such as by choosing the set of points \( \mathbf{p} \) with the largest movement distance. The size of this Gaussian set \( G \) can be as minimal as a single Gaussian.

Utilizing the explicit representation of GS, we calculate the centroid \( \mathbf{c} \) of the points in \( G \) and trace its movement trajectory. We employ a simple linear model for trajectory analysis, although more complex models are feasible. Principal Component Analysis (PCA) is applied to determine the primary movement direction, denoted as \( \mathbf{d} \). The positions of the Gaussian (means) \( \mu \) are then projected onto this direction \( \mathbf{d} \) at each timestep \( t \), as shown in Eq.~\ref{eq:proj}:

\begin{equation}
   \mathbf{proj}_{\mathbf{d}}(\mu_t) = \frac{(\mu_t - \mathbf{c}) \cdot \mathbf{d}}{\|\mathbf{d}\|}. 
  \label{eq:proj}
\end{equation}

This projection enables us to define the start and end points, \( \mathbf{s} \) and \( \mathbf{e} \), along the movement direction. Subsequently, the distances of all points from the start point \( \mathbf{s} \) are normalized to a range between 0 and 1, resulting in our control signal \( \sigma \), as expressed in Eq.~\ref{eq:norm}:

\begin{equation}
 \sigma(\mu_t) = \frac{\mathbf{proj}_{\mathbf{d}}(\mu_t) - \mathbf{proj}_{\mathbf{d}}(\mathbf{s})}{\mathbf{proj}_{\mathbf{d}}(\mathbf{e}) - \mathbf{proj}_{\mathbf{d}}(\mathbf{s})}.  
  \label{eq:norm}
\end{equation}

This process culminates in the control signal \( \sigma \), enabling dynamic scene manipulation.


\subsubsection{Control Signal Re-Alignment}

After obtaining the control signal \( \sigma \), the next crucial step is its integration into the network to facilitate manipulation using these signals. This is accomplished by developing a unique network \( N^c_i \) for each control signal, designed to output the corresponding offset \( \Delta \) for each Gaussian attribute, as determined in the dynamic modeling stage. Let $\mu_i$, $\mathbf{C}_i$, $\mathbf{R}_i$, and $\mathbf{S}_i$ denote the mean, rotation, and scaling of each Gaussian, respectively. The control network \( N^c_i \) modifies these attributes in response to the control signal:
\begin{equation}
  N^c_i(\sigma) = (\Delta \mu_i, \Delta \mathbf{C}_i, \Delta  \mathbf{R_i}, \Delta  \mathbf{S_i})
  \label{eq:L3}
\end{equation}

In this phase, the focus is solely on training these control signal networks \( N_i \), while keeping all other learnable parameters \( \Theta \)  fixed.

Upon achieving reliable estimates for the attribute offsets \( \Delta \mu_i, \Delta \mathbf{C}_i, \Delta  \mathbf{R_i}, \Delta  \mathbf{S_i} \) of each Gaussian, we move towards the end-to-end fine-tuning of all learnable parameters. This all-encompassing fine-tuning is crucial for completing our controllable GS model. The final model (represented by $f$) is formulated as:
\begin{equation}
  f(\Theta; \mu_i + \Delta \mu_i, \mathbf{C}_i + \Delta \mathbf{C}_i,  \mathbf{R_i} + \Delta  \mathbf{R_i},  \mathbf{S_i} + \Delta  \mathbf{S_i})
  \label{eq:L3}
\end{equation}

This final step guarantees precise and effective control over dynamic scene renderings.

\section{Experiments}

\begin{figure}[ht]
  \centering
  \includegraphics[width=0.44\textwidth]{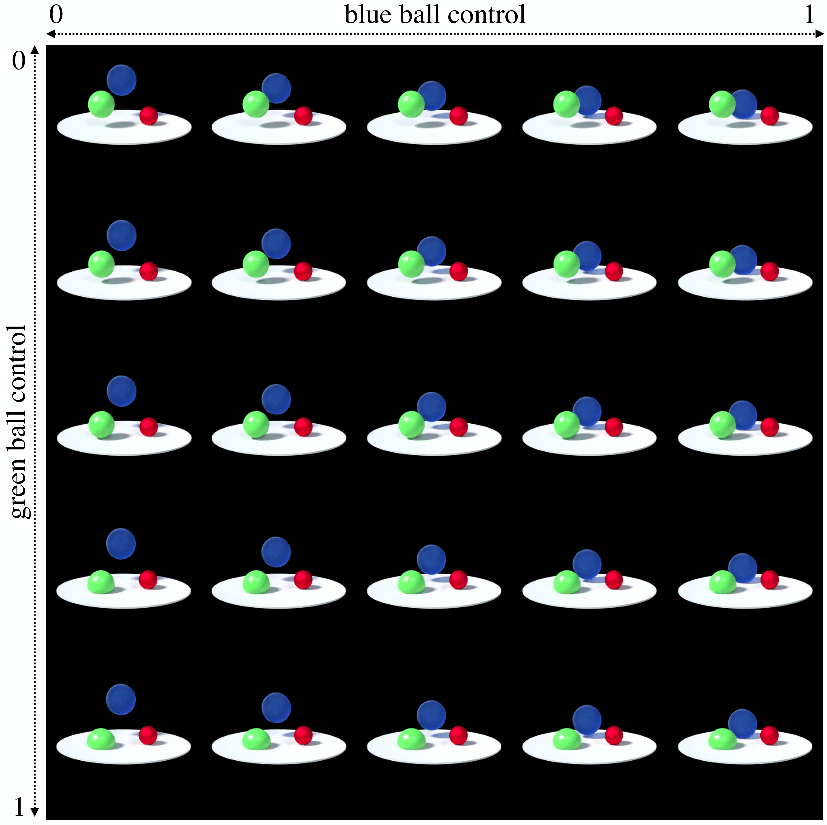}
  \caption{Control blue ball and green ball separately.}
  \label{ctrl:ball}
\end{figure}

\begin{figure}[ht]
  \centering
  \includegraphics[width=0.44\textwidth]{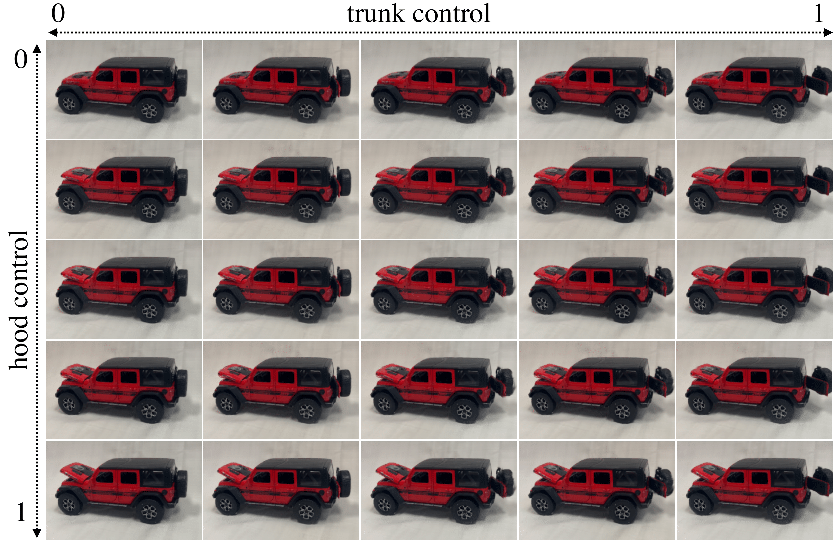}
  \caption{Opening the hood and the trunk of the toy car.}
  \label{ctrl:car}
\end{figure}

In this section, we present the experiments conducted to demonstrate the effectiveness of our method in dynamic and controllable scenarios.

\begin{figure}[ht]
\centering
\begin{subfigure}{0.11\textwidth}
  \centering
  \includegraphics[width=0.9\linewidth]{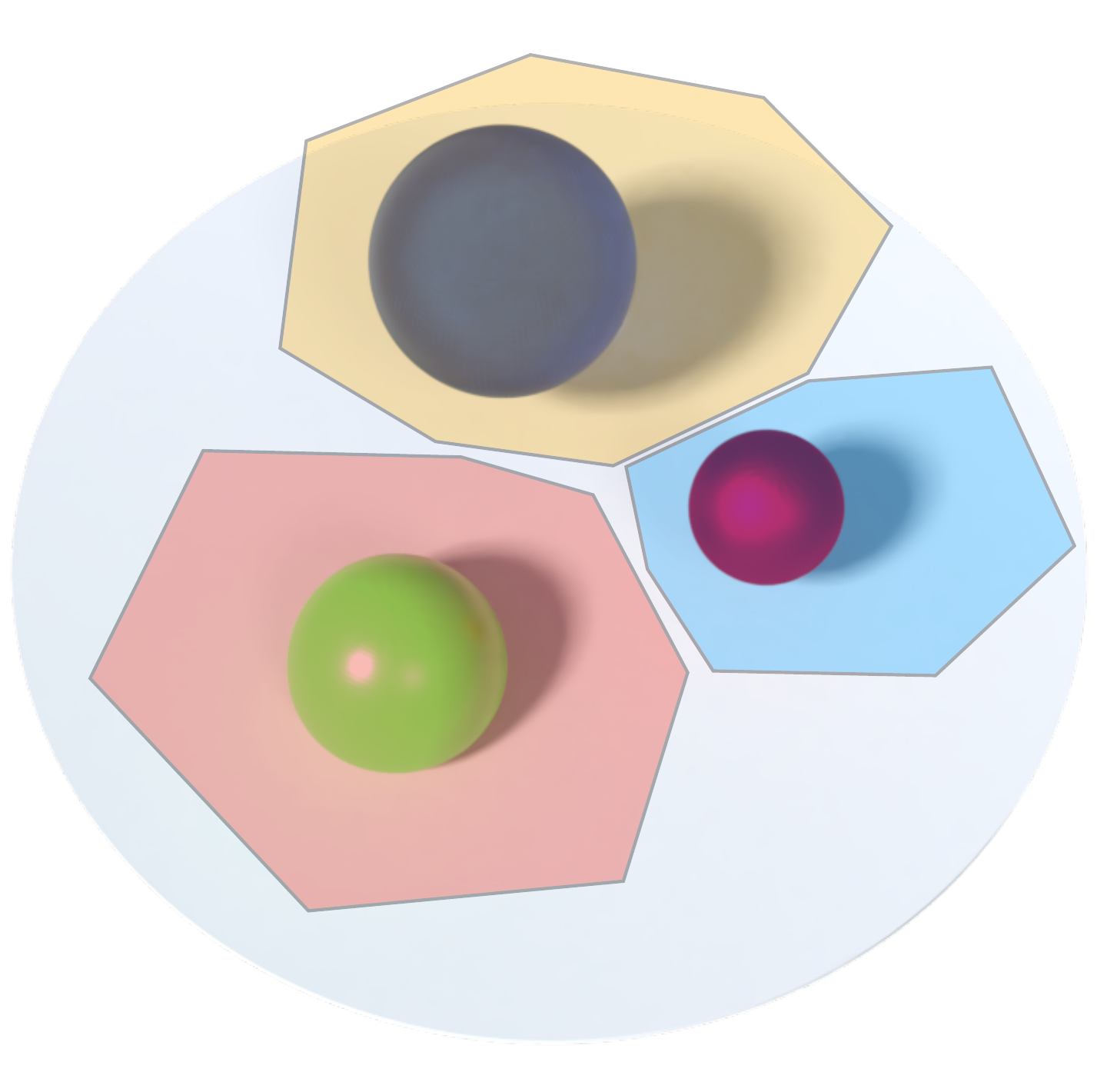}
  \caption{}
  \label{fig:ball_mask}
\end{subfigure}%
\begin{subfigure}{0.11\textwidth}
  \centering
  \includegraphics[width=0.9\linewidth]{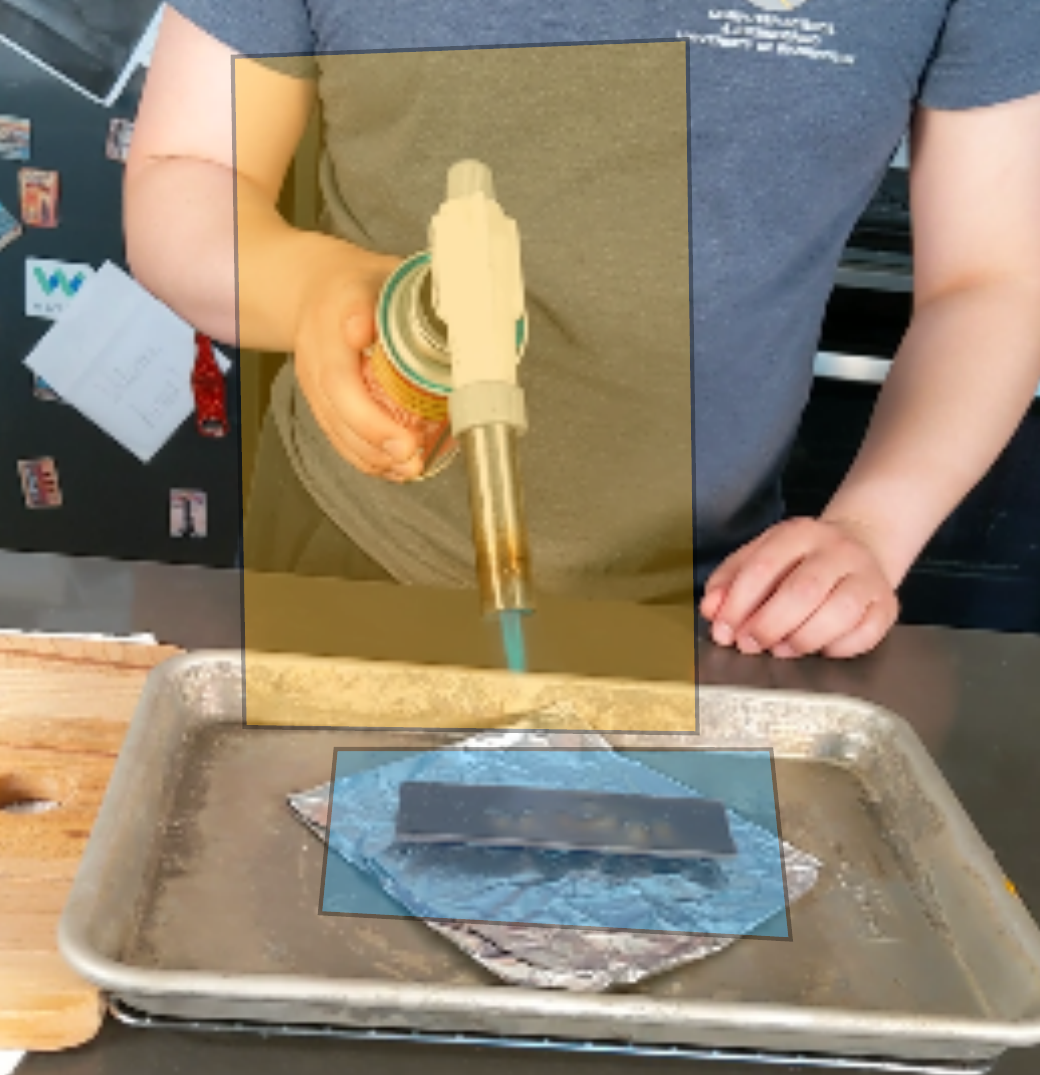}
  \caption{}
  \label{fig:torchchoco_mask}
\end{subfigure}
\begin{subfigure}{0.11\textwidth}
  \centering
  \includegraphics[width=0.9\linewidth]{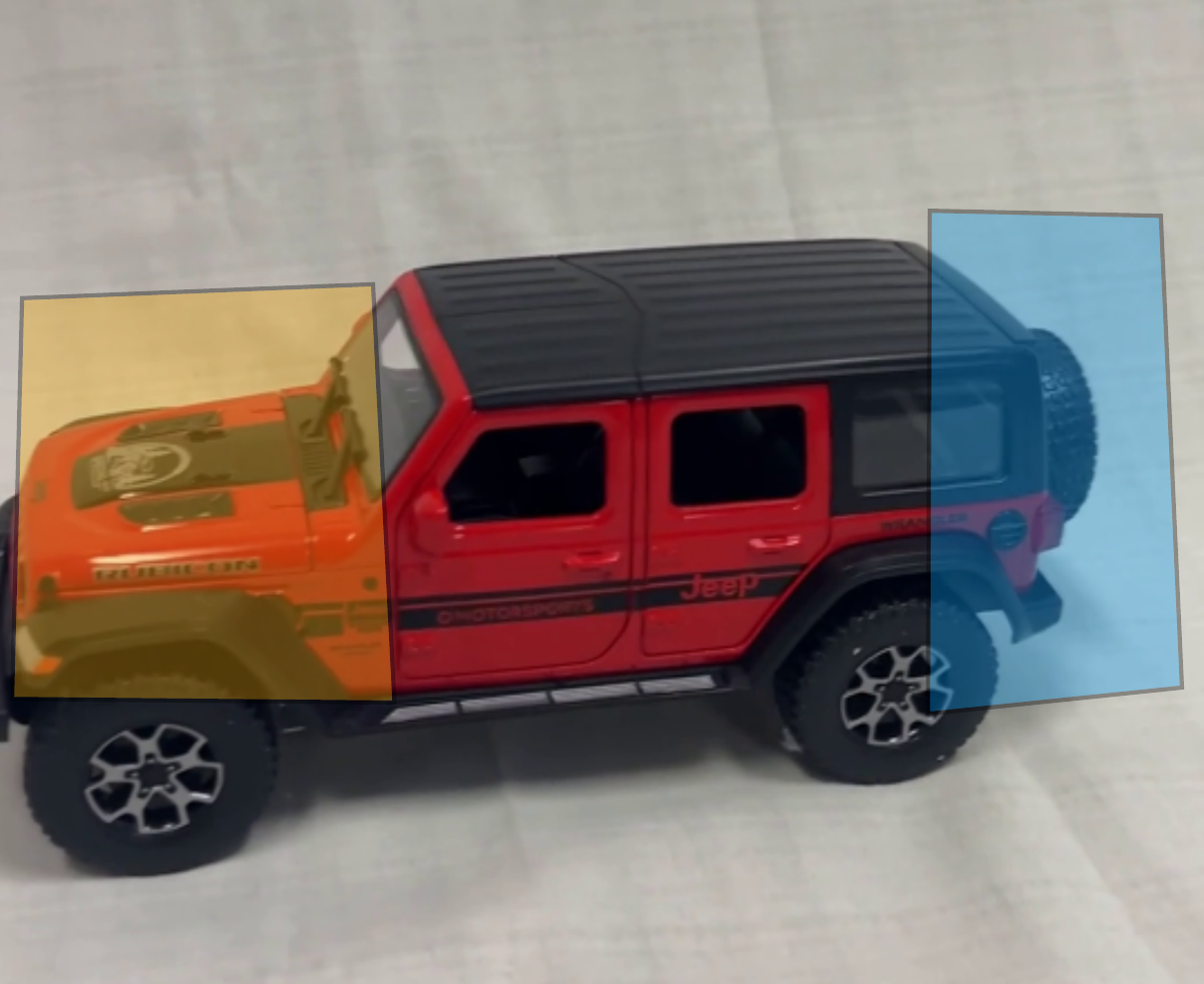}
  \caption{}
  \label{fig:car_mask}
\end{subfigure}
\begin{subfigure}{0.11\textwidth}
  \centering
  \includegraphics[width=0.9\linewidth]{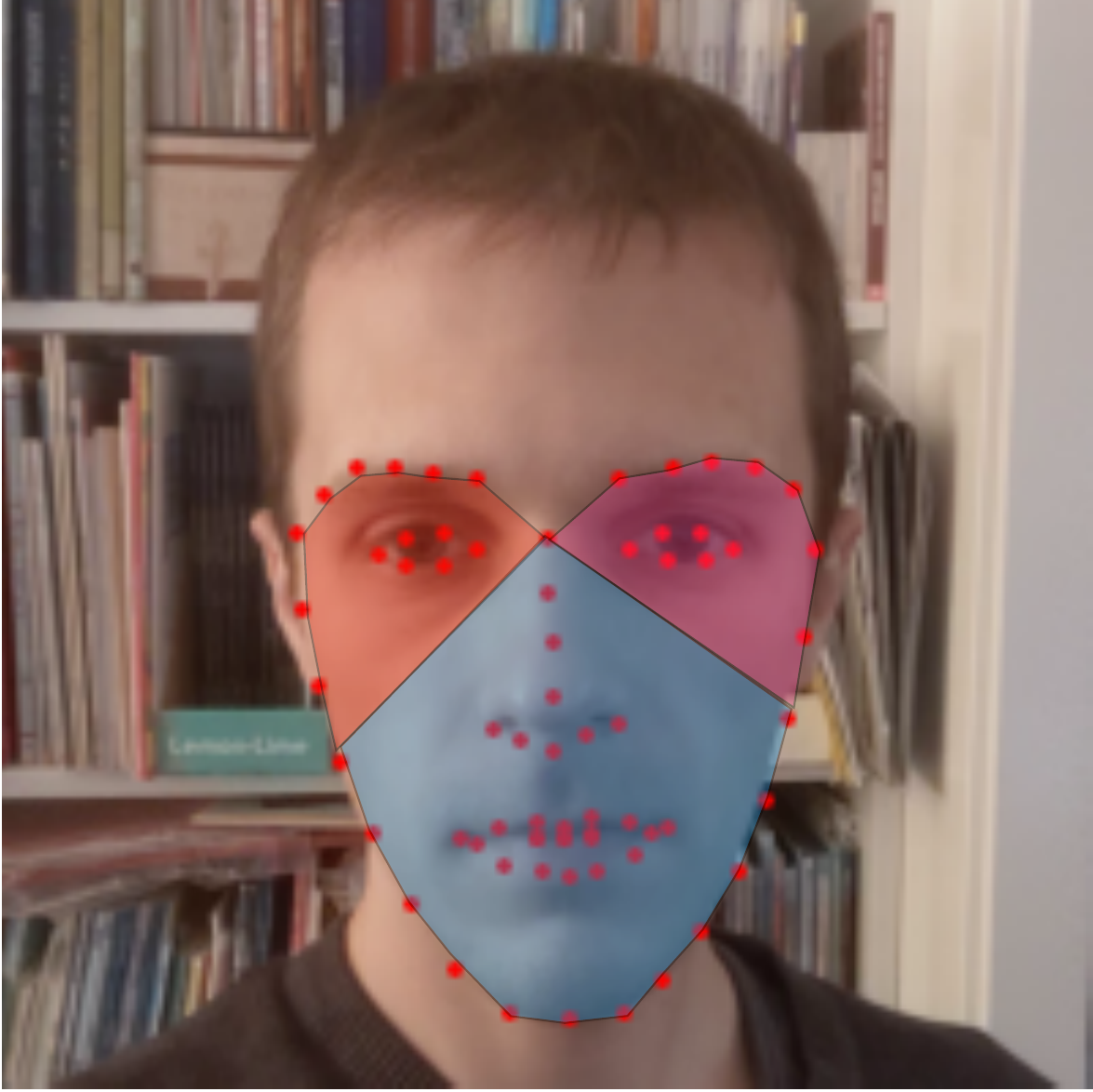}
  \caption{}
  \label{fig:face_mask}
\end{subfigure}
\caption{2D mask labeling. (a)-(c) Manually labeling a single frame. (d) Automatic labeling using facial key-point detection.}
\label{fig:mask}
\end{figure}

\begin{table}[htb]
\centering
\caption{Quantitative results on synthetic dynamic scenes. We color code each row as \colorbox{red!25}{best}, \colorbox{orange!25}{second best}, and \colorbox{yellow!25}{third best}.}
\resizebox{0.7\columnwidth}{!}{%

\begin{tabular}{lccc}
\toprule
Method                                 & PSNR$\uparrow$ & SSIM$\uparrow$ & LPIPS$\downarrow$ (100x) \\ \midrule
NeRF\cite{mildenhall2021nerf}          & 18.98          & 0.870           &18.25              \\
DirectVoxGo\cite{sun2022direct}              & 18.64          & 0.853           &16.88              \\
Plenoxels\cite{fridovich2022plenoxels} & 20.24          & 0.868          &16.00              \\
T-NeRF\cite{pumarola2021d}             & 29.50          & 0.951           & 7.88              \\
D-NeRF\cite{pumarola2021d}             & 30.44          & 0.952           & 6.63              \\
TiNeuVox-S\cite{tineuvox}              & 30.75          & 0.955           & 6.63              \\
TiNeuVox-B\cite{tineuvox}              & 32.67          & 0.972           & 4.25              \\ 

3D GS~\cite{kerbl20233d}         & 23.07          & 0.928           & 8.22                  \\
\midrule
Ours              & \cellcolor{red!25}37.90          & \cellcolor{orange!25}0.983           & \cellcolor{red!25}1.74                  \\
Ours, w/o $\mathcal{L}^{\text{norm}}$             & 37.41          & \cellcolor{red!25}0.984           & \cellcolor{yellow!25}1.70                  \\
Ours, w/o $\mathcal{L}^{\text{diff}}$              & \cellcolor{yellow!25}37.68          & \cellcolor{yellow!25}0.982           & 1.65                  \\
Ours, w/o $\mathcal{L}^{\text{rigid}}$             & \cellcolor{orange!25}37.75          & 0.981           & \cellcolor{orange!25}1.71                  \\
\bottomrule
\end{tabular}%
}
\label{tab:synth}
\end{table}


\begin{table}[ht]
\centering
\caption{Quantitative results on real dynamic scenes. We color code each row as \colorbox{red!25}{best}, \colorbox{orange!25}{second best}, and \colorbox{yellow!25}{third best}.}
\resizebox{0.7\columnwidth}{!}{%
\begin{tabular}{lccc}
\toprule
Method            & PSNR$\uparrow$ & SSIM$\uparrow$ & LPIPS$\downarrow$ (100x)\\ \midrule
NeRF\cite{mildenhall2021nerf}          &  22.3 &	0.807 &	43.3                  \\
NV\cite{lombardi2019neural}                &   26.3 &	0.910 &	20.9              \\
NSFF\cite{li2021neural}              &   25.7 &	0.881 &	24.8                         \\
Nerfies\cite{park2021nerfies}       &  29.3 &	\cellcolor{yellow!25}0.948 &	\cellcolor{yellow!25}17.6                \\
HyperNeRF\cite{park2021hypernerf}   & \cellcolor{orange!25}29.8 &	\cellcolor{red!25}0.954 &	\cellcolor{orange!25}17.2                \\
TiNeuVox-S\cite{tineuvox}              &  23.6 &	0.690 &	54.5                 \\
TiNeuVox-B\cite{tineuvox}              &    28.0 &	0.752 &	45.1             \\
3D GS~\cite{kerbl20233d}                               &   22.1 &	0.724 &	43.8                             \\
\midrule
Ours                            &   \cellcolor{yellow!25}29.6 &	\cellcolor{orange!25}0.950 &	\cellcolor{red!25}17.1                         \\
Ours, w/o $\mathcal{L}^{\text{norm}}$             & 29.1         & 0.905           & 20.1                  \\
Ours, w/o $\mathcal{L}^{\text{diff}}$              & \cellcolor{red!25}29.8          & 0.912           & 19.8                  \\
Ours, w/o $\mathcal{L}^{\text{rigid}}$             & 29.4          & 0.920            & 21.3                  \\

\bottomrule
\end{tabular}%
}
\label{tab:real}
\end{table}

\subsection{Datasets}


To evaluate our dynamic model, experiments were conducted on two categories of dynamic scenes: synthetic and real. Additionally, the performance of our controllable model was assessed on a synthetic scene, real face scene, real dynamic scene, and a self-captured toy car scene. \\
\textbf{Synthetic Scenes.} We employed the $360^{\circ}$ dynamic synthetic dataset introduced by \cite{pumarola2021d}, comprising 8 animated objects with complex geometries and non-Lambertian materials. Each scene in this dataset includes $50$ to $200$ training images and $20$ test images, all at an $800 \times 800$ resolution.\\
\textbf{Real Scenes.} Four topologically diverse scenes from \cite{park2021hypernerf} (torchocolate, cut-lemon, chickchicken, and hand) were used. These scenes were captured using a rig consisting of two Google Pixel 3 phones mounted approximately 16cm apart on a pole.\\
\textbf{Real Face Scene.} For controllable model testing, we utilized a real face scene from \cite{kania2022conerf} (involving actions like closing/opening the eyes/mouth). This scene was captured with either a Google Pixel 3a or an Apple iPhone 13 Pro. Unlike CoNeRF, which requires pre-defined control signals and masks, our method achieves comparable controllability without such prerequisites.\\
\textbf{Toy Car Scene.} We also ran experiments on two self-captured Toy Car scenes. The capture process involved manually opening the regions of control (doors, hood, trunk, etc.) and recording one video per transition. Once the transitions were completed, we stitched the individual videos together to create a single cohesive dynamic scene. This scene was captured with an Apple iPhone 13 Pro.

\begin{figure*}[!ht]
     \centering
     \begin{subfigure}[b]{0.15\textwidth}
         \centering
         \includegraphics[width=0.5\textwidth, height=2.03cm,keepaspectratio]{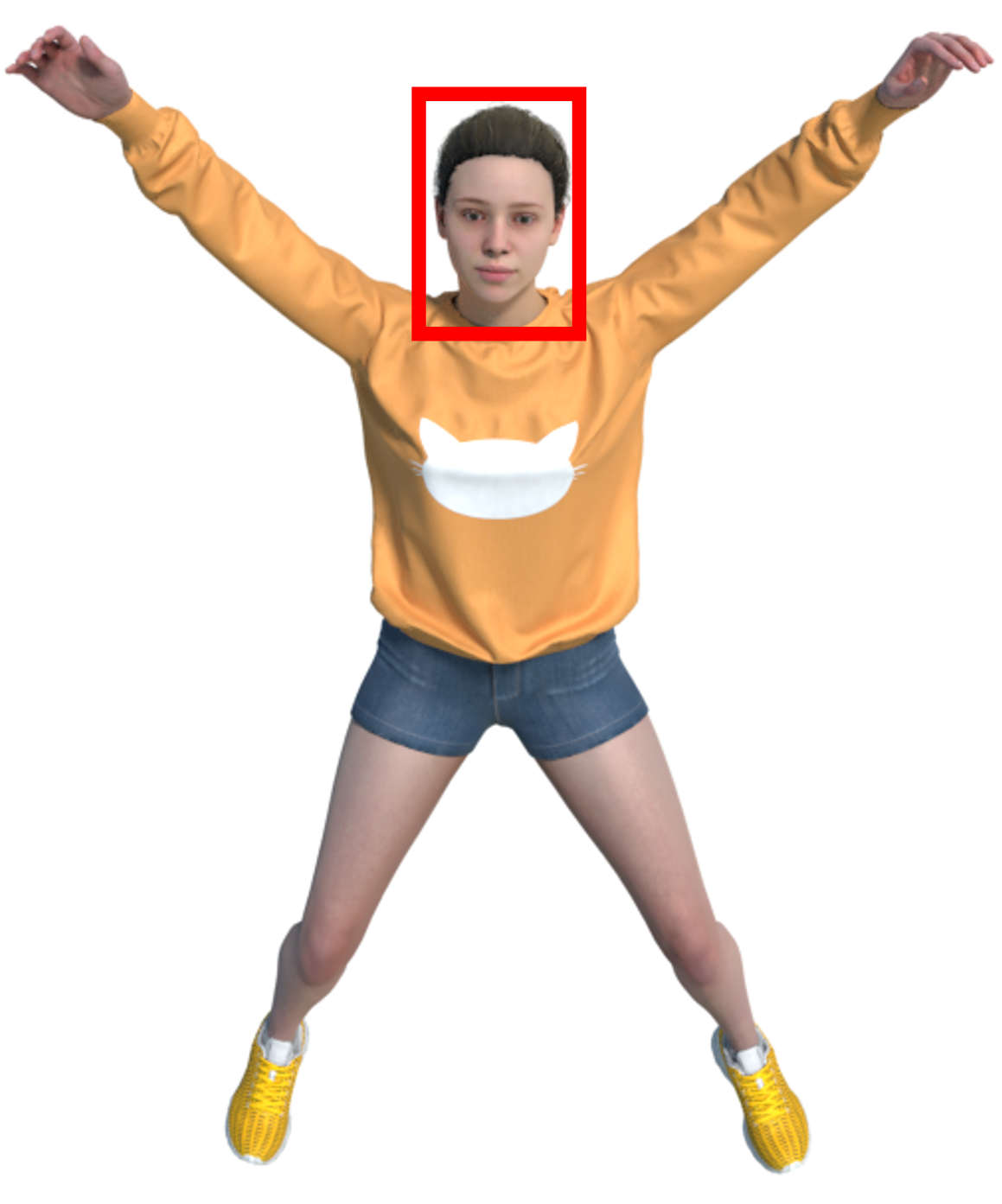}
         \caption*{\textbf{Jumping Jacks}}
         \label{fig:chicken_box}
     \end{subfigure}     
     \setcounter{subfigure}{0}
     \begin{subfigure}[b]{0.15\textwidth}
         \centering
         \includegraphics[width=0.5\textwidth]{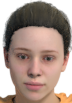}
         \caption{Ground Truth}
         \label{fig:chicken_gt}
     \end{subfigure}
     \begin{subfigure}[b]{0.15\textwidth}
         \centering
         \includegraphics[width=0.5\textwidth]{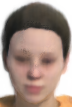}
         \caption{D-NeRF \cite{pumarola2021d}}
         \label{fig:chicken_hypernerf}
     \end{subfigure}
    \begin{subfigure}[b]{0.15\textwidth}
         \centering
         \includegraphics[width=0.5\textwidth]{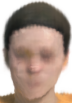}
         \caption{TiNeuVox \cite{tineuvox}}
         \label{fig:chicken_tineuvox}
     \end{subfigure}
    \begin{subfigure}[b]{0.15\textwidth}
         \centering
         \includegraphics[width=0.5\textwidth]{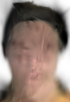}
         \caption{3D-GS \cite{kerbl20233d}}
         \label{fig:chicken_ours1}
     \end{subfigure}
    \begin{subfigure}[b]{0.15\textwidth}
         \centering
         \includegraphics[width=0.5\textwidth]{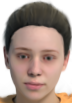}
         \caption{Ours}
         \label{fig:chicken_ours2}
     \end{subfigure}    
     \begin{subfigure}[b]{0.15\textwidth}
         \centering
         \includegraphics[width=0.5\textwidth,height=2.03cm,keepaspectratio]{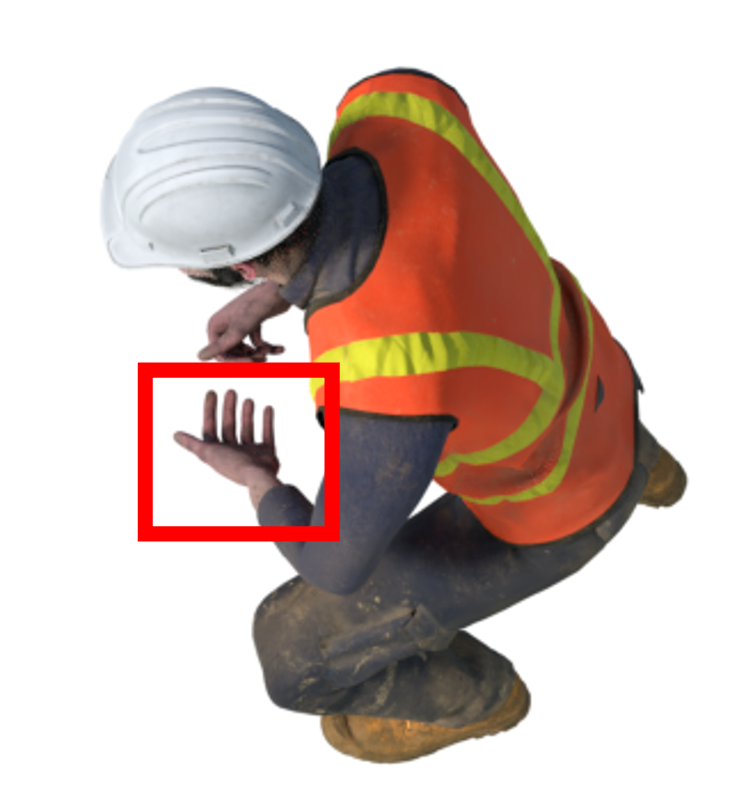}
         \caption*{\textbf{Standup}}
         \label{fig:americano_box}
     \end{subfigure}     
     \begin{subfigure}[b]{0.15\textwidth}
         \centering
         \includegraphics[width=0.5\textwidth]{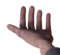}
         \caption{Ground Truth}
         \label{fig:americano_gt}
     \end{subfigure}
     \begin{subfigure}[b]{0.15\textwidth}
         \centering
         \includegraphics[width=0.5\textwidth]{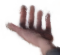}
         \caption{D-NeRF \cite{pumarola2021d}}
         \label{fig:americano_hypernerf}
     \end{subfigure}
    \begin{subfigure}[b]{0.15\textwidth}
         \centering
         \includegraphics[width=0.5\textwidth]{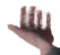}
         \caption{TiNeuVox \cite{tineuvox}}
         \label{fig:americano_tineuvox}
     \end{subfigure}
    \begin{subfigure}[b]{0.15\textwidth}
         \centering
         \includegraphics[width=0.5\textwidth]{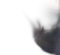}
         \caption{3D-GS \cite{kerbl20233d}}
         \label{fig:americano_ours1}
     \end{subfigure}
    \begin{subfigure}[b]{0.15\textwidth}
         \centering
         \includegraphics[width=0.5\textwidth]{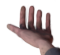}
         \caption{Ours}
         \label{fig:americano_ours2}
     \end{subfigure}    

        \caption{
        Qualitative results on synthetic dynamic scenes.
        We compare our Dynamic 3D-GS method (Ours) with the ground truth, D-NeRF, TiNeuVox, and the static 3D-GS method.}
        \label{fig:qual_real}
\end{figure*}

\begin{figure*}[!ht]
     \centering
     \begin{subfigure}[b]{0.15\textwidth}
         \centering
         \includegraphics[width=0.7\textwidth, height=2.03cm,keepaspectratio]{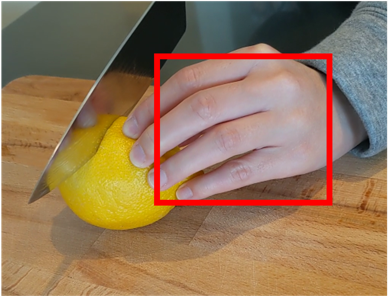}
         \caption*{\textbf{Cut Lemon}}
         \label{fig:chicken_box}
     \end{subfigure}     
     \setcounter{subfigure}{0}
     \begin{subfigure}[b]{0.15\textwidth}
         \centering
         \includegraphics[width=0.7\textwidth]{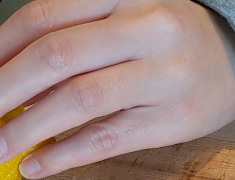}
         \caption{Ground Truth}
         \label{fig:chicken_gt}
     \end{subfigure}
     \begin{subfigure}[b]{0.15\textwidth}
         \centering
         \includegraphics[width=0.7\textwidth]{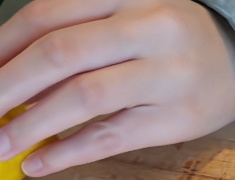}
         \caption{HyperNeRF \cite{park2021hypernerf}}
         \label{fig:chicken_hypernerf}
     \end{subfigure}
    \begin{subfigure}[b]{0.15\textwidth}
         \centering
         \includegraphics[width=0.7\textwidth]{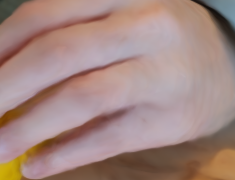}
         \caption{TiNeuVox \cite{tineuvox}}
         \label{fig:chicken_tineuvox}
     \end{subfigure}
    \begin{subfigure}[b]{0.15\textwidth}
         \centering
         \includegraphics[width=0.7\textwidth]{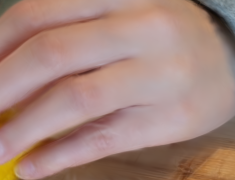}
         \caption{3D-GS \cite{kerbl20233d}}
         \label{fig:chicken_ours1}
     \end{subfigure}
    \begin{subfigure}[b]{0.15\textwidth}
         \centering
         \includegraphics[width=0.7\textwidth]{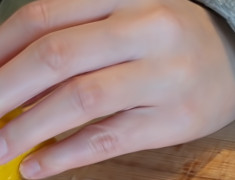}
         \caption{Ours}
         \label{fig:chicken_ours2}
     \end{subfigure}    
     \begin{subfigure}[b]{0.15\textwidth}
         \centering
         \includegraphics[width=0.5\textwidth,height=2.03cm,keepaspectratio]{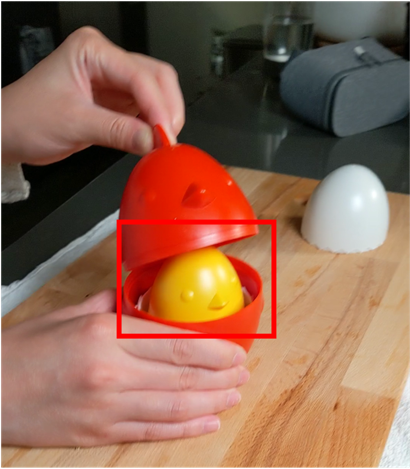}
         \caption*{\textbf{Chick Chicken}}
         \label{fig:americano_box}
     \end{subfigure}     
     \begin{subfigure}[b]{0.15\textwidth}
         \centering
         \includegraphics[width=0.7\textwidth]{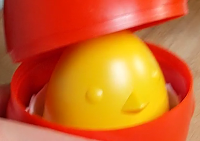}
         \caption{Ground Truth}
         \label{fig:americano_gt}
     \end{subfigure}
     \begin{subfigure}[b]{0.15\textwidth}
         \centering
         \includegraphics[width=0.7\textwidth]{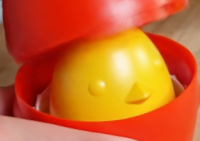}
         \caption{HyperNeRF \cite{park2021hypernerf}}
         \label{fig:americano_hypernerf}
     \end{subfigure}
    \begin{subfigure}[b]{0.15\textwidth}
         \centering
         \includegraphics[width=0.7\textwidth]{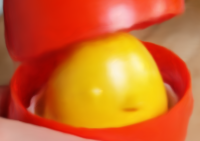}
         \caption{TiNeuVox \cite{tineuvox}}
         \label{fig:americano_tineuvox}
     \end{subfigure}
    \begin{subfigure}[b]{0.15\textwidth}
         \centering
         \includegraphics[width=0.7\textwidth]{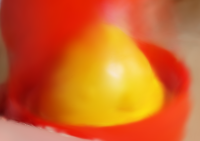}
         \caption{3D-GS \cite{kerbl20233d}}
         \label{fig:americano_ours1}
     \end{subfigure}
    \begin{subfigure}[b]{0.15\textwidth}
         \centering
         \includegraphics[width=0.7\textwidth]{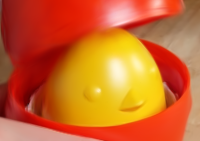}
         \caption{Ours}
         \label{fig:americano_ours2}
     \end{subfigure}    

        \caption{
        Qualitative results on real dynamic scenes.
        We compare our Dynamic 3D-GS method (Ours) with the ground truth, HyperNeRF, TiNeuVox, and the static 3D-GS method. For Cut Lemon, our method models the knuckles on the hand better than others. As for Chick Chicken, we reconstruct more fine details (red edges).}
        \label{fig:qual_real}
\end{figure*}

\subsection{Implementation Details}

For the training of the dynamic component, we adopted the differential Gaussian rasterization technique from 3D GS \cite{kerbl20233d}, and implemented the additional network components using PyTorch \cite{paszke2019pytorch}. The Gaussians were initialized either using SfM results or by randomly selecting $N$ points (where $N=10k$ in our experiments) within the scene box. The initial phase of 3k iterations does not involve learning any deformation field; this phase is akin to the training process of 3D GS, aiding in the convergence of the learning process. Following this, we jointly train the 3D Gaussian attributes and the deformation network for a total of 50k iterations. The learning rate for each Gaussian attribute is kept consistent with that used in 3D GS, as detailed in \cite{kerbl20233d}, while the learning rate for the deformation network is set to exponentially decay from $1e-3$ to $1e-6$.

In the controllable part, the learning rate is set to 1 for the initial 1k iterations during the 3D Mask Generation phase. During the Control Signal Re-Alignment phase, we apply an exponential decay of the learning rate from $1e-2$ to $1e-4$ over 5k iterations, specifically for training the control signal networks. For the final end-to-end finetuning, the learning rate is set to $1e-6$, and the process is run for an additional 5k iterations. Optimization throughout these processes is performed using the Adam optimizer \cite{kingma2014adam} with a $\beta$ value range of $(0.9, 0.999)$. All experiments were conducted using single 80GB NVIDIA A100 GPUs.

\subsection{Results}
We show the qualitative and quantitative results in this section to demonstrate the effectiveness of our method. We use Peak Signal-to-Noise Ratio (PSNR) \cite{hore2010image} in decibels (dB), the Structural Similarity Index (SSIM) \cite{wang2004image, odena2017conditional} and the Learned Perceptual Image Patch Similarity (LPIPS) \cite{zhang2018unreasonable} as evaluation metrics. All detailed results for each scene can be found in the supplementary material. 

We first show our dynamic GS modeling part. We compare our method with existing works using dynamic synthetic scenes from \cite{pumarola2021d} on the novel view synthesis task.  We report the quantitative results in Table~\ref{tab:synth} and we can see that our method can achieve much better performance than existing methods. The qualitative results are shown in Fig.~\ref{fig:qual_real} and we can see that our method has better face and hand details. For the real scenes from \cite{park2021hypernerf}, we run interpolation experiments as in~\cite{park2021hypernerf} instead of the novel view synthesis task because of the rendering pose problem mentioned in~\cite{yang2023deformable}. We show our results in Table~\ref{tab:real} and Fig.~\ref{fig:qual_real}. We can see that our method can capture better details on complex real dynamic scenes. We also perform ablation experiments on the regularizations we utilize as shown in Table~\ref{tab:synth} \& \ref{tab:real}. To better illustrate the role of the regularizations, we also visualize the trajectories in Fig.~\ref{fig:ablation}. 

For the controllable GS, we use four datasets (bouncingball, torchocolate, face and car) and first fit a dynamic GS on them. Then we obtain the labels as shown in Fig.~\ref{fig:mask}. For the eye scene, we manually select the point set as the control points, and for other scenes, we get the point sets automatically from the point movement as mentioned before. We also visualize the control results to demonstrate our method's performance.

\begin{figure}[ht]
  \centering
  \includegraphics[width=0.44\textwidth]{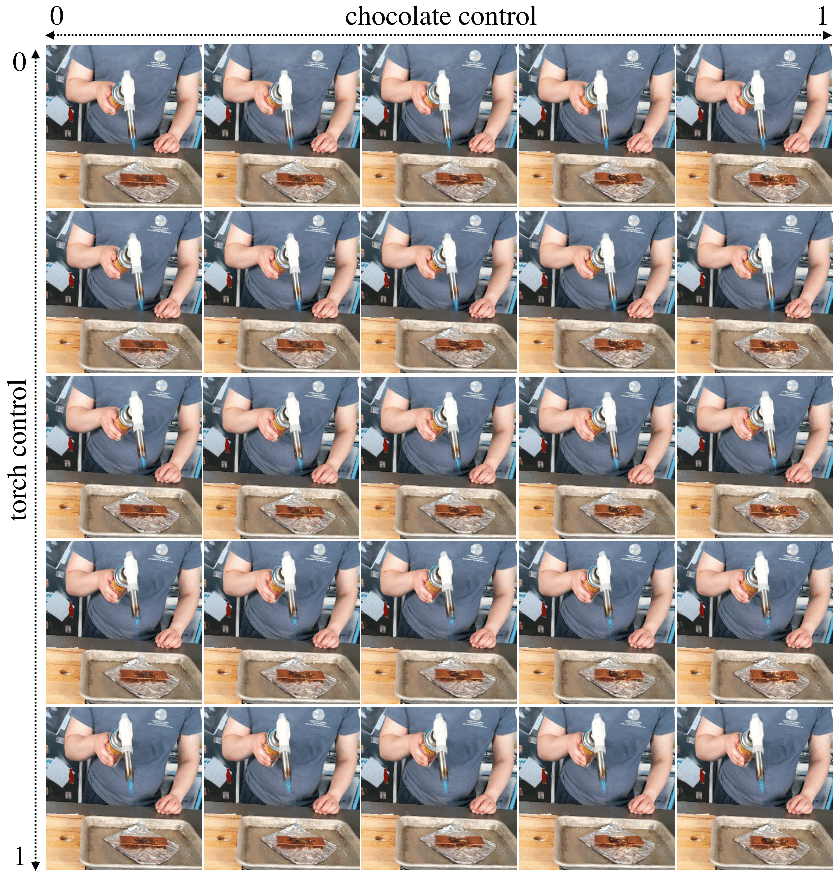}
  \caption{Controlling the blowtorch and the melting chocolate separately.}
  \label{ctrl:torch}
\vspace{-0.25cm}
\end{figure}

\begin{figure}[ht]
  \centering
  \includegraphics[width=0.44\textwidth]{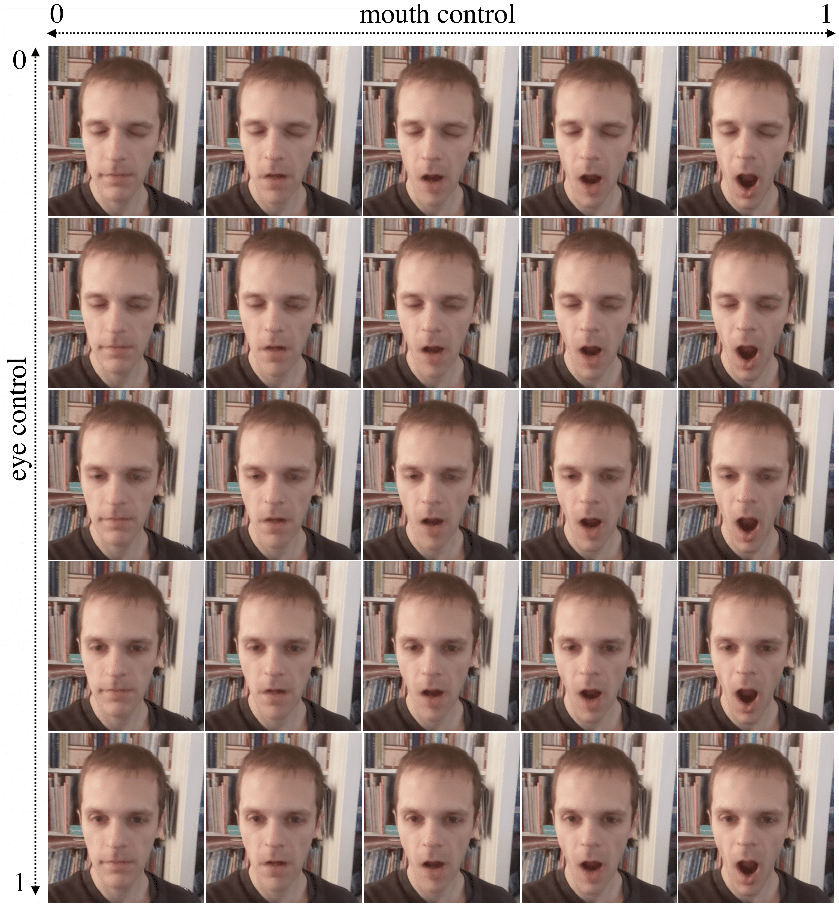}
  \caption{Controlling the eyes and the mouth separately.}
  \label{ctrl:eye_mouth}
\vspace{-0.25cm}
\end{figure}

\begin{figure}[ht]
  \centering
  \includegraphics[width=0.44\textwidth]{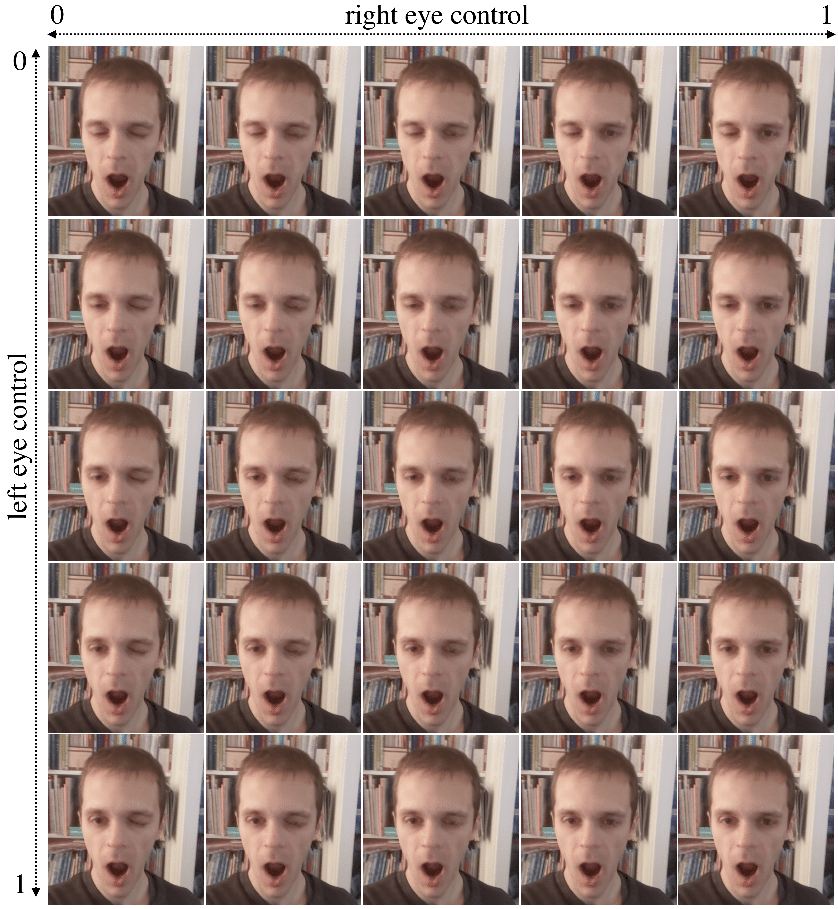}
  \caption{Controlling the left eye and right eye separately.}
  \label{ctrl:eye_eye}
\vspace{-0.35cm}
\end{figure}

\section{Conclusion}
We presented Controllable Gaussian Splatting named CoGS, a novel method for dynamic scene manipulation. It overcomes the limitations of NeRFs and similar neural methods by using an explicit representation that enables real-time, controllable manipulation of dynamic scenes. Our approach, validated through extensive experiments, shows superior performance in visual fidelity and manipulation capabilities compared to existing techniques. The explicit nature of CoGS not only enhances efficiency in rendering but also simplifies scene element manipulation. It has the potential to democratize 3D deformable content creation using commodity hardware, making it more accessible and feasible for a broader range of users and applications.

Our method is not without limitations. CoGS faces challenges with shiny or intricately lit objects, common in GS pipelines. Dynamic modeling may struggle with non-rigid deformation and large-scale movements in monocular settings. Limitations may also arise from controllable signal extraction and re-alignment, with the current PCA method potentially struggling with highly complex movements. Addressing these limitations will be the focus of future work. 

\section*{Acknowledgements}
This research was supported partially by Fujitsu.



{
    \small
    \bibliographystyle{ieeenat_fullname}
    \bibliography{main}
}

\clearpage
\setcounter{page}{1}
\maketitlesupplementary


\section{Overview}
This supplementary material offers comprehensive quantitative data and further qualitative insights, highlighting the advantages of our newly developed Dynamic 3D Gaussian Splatting (GS) and Controllable GS methods. In addition, we have included illustrative videos on the attached webpage, providing a dynamic visual representation of our methods in action.

\section{Per-Scene Quantitative Results}
For completeness, we provide detailed per-scene quantitative results for reconstruction quality metrics, including PSNR, SSIM, and LPIPS. These are presented for both synthetic (\cref{tab:synth2}) and real (\cref{tab:real2}) dynamic scenes. This extension to Tables~\textcolor{red}{1} and~\textcolor{red}{2} from the main paper offers a more nuanced view, as it disaggregates the average performance metrics across different scenes. Our analysis reveals that our Dynamic GS method exhibits superior performance in synthetic scene datasets while achieving comparable results in real scenes. This difference in performance might be attributed to the challenges inherent in modeling the movement of Gaussians using a single camera setup.

\section{More Qualitative Results}
Additional qualitative results can be found on the project's website (\url{https://cogs2024.github.io}). For optimal viewing, please open the link using the Chrome browser.

\begin{table*}[h]
\centering
\caption{Per-scene quantitative results on synthetic dynamic scenes. We color code each row as \colorbox{red!25}{best}, \colorbox{orange!25}{second best}, and \colorbox{yellow!25}{third best}.}
\resizebox{\textwidth}{!}{%
\begin{tabular}{lcccccccccccc}
\toprule
                                 & \multicolumn{3}{c}{Hell Warrior}                              & \multicolumn{3}{c}{Mutant}                                     & \multicolumn{3}{c}{Hook}                                          & \multicolumn{3}{c}{Bouncing Balls}                                     \\
\cmidrule(lr){2-4} \cmidrule(lr){5-7} \cmidrule(lr){8-10} \cmidrule(lr){11-13}
Method                           & PSNR$\uparrow$           & SSIM$\uparrow$ & LPIPS$\downarrow$ & PSNR$\uparrow$            & SSIM$\uparrow$ & LPIPS$\downarrow$ & PSNR$\uparrow$               & SSIM$\uparrow$ & LPIPS$\downarrow$ & PSNR$\uparrow$                    & SSIM$\uparrow$ & LPIPS$\downarrow$ \\ \midrule
NeRF\cite{mildenhall2021nerf}    & 13.52                    & 0.8100           & 0.2500              & 20.31                     & 0.9100           & 0.0900              & 16.65                        & 0.8400           & 0.1900              & 20.26                             & 0.9100           & 0.2000              \\
DirectVoxGo\cite{sun2022direct}        & 13.51                    & 0.7500           & 0.2500              & 19.45                     & 0.8900           & 0.1200              & 16.16                        & 0.8000           & 0.2100              & 20.20                             & 0.8700           & 0.2200              \\
Plenoxels\cite{fridovich2022plenoxels}& 15.19               & 0.7800           & 0.2700              & 21.44                     & 0.9100           & 0.0900              & 17.90                        & 0.8100           & 0.2100              & 21.30                             & 0.8900           & 0.1800              \\
T-NeRF\cite{pumarola2021d}       & 23.19                    & 0.9300           & 0.0800              & 30.56                     & 0.9600           & 0.0400              & 27.21                        & 0.9400           & \cellcolor{yellow!25}0.0600              & 37.81                             & 0.9800           & 0.1200              \\
D-NeRF\cite{pumarola2021d}       & 25.10                    & 0.9500           & \cellcolor{orange!25}0.0600              & \cellcolor{yellow!25}31.29                     & \cellcolor{yellow!25}0.9700           & \cellcolor{orange!25}0.0200              & 29.25                        & \cellcolor{yellow!25}0.9600           & 0.1100              & 38.93                             & 0.9800           & 0.1000              \\
TiNeuVox-S\cite{tineuvox}        & 27.00                    & \cellcolor{yellow!25}0.9500           & 0.0900              & 31.09                     & 0.9600           & 0.0500              & \cellcolor{yellow!25}29.30                        & 0.9500           & 0.0700              & \cellcolor{yellow!25}39.05                             & \cellcolor{yellow!25}0.9900           & \cellcolor{yellow!25}0.0600              \\
TiNeuVox-B\cite{tineuvox}        & \cellcolor{yellow!25}28.17                   & \cellcolor{orange!25}0.9700           & \cellcolor{yellow!25}0.0700              & \cellcolor{orange!25}33.61                     & \cellcolor{orange!25}0.9800           & \cellcolor{yellow!25}0.0300              & \cellcolor{orange!25}31.45                        & \cellcolor{orange!25}0.9700           & \cellcolor{orange!25}0.0500              & \cellcolor{orange!25}40.73                             & \cellcolor{orange!25}0.9900           & \cellcolor{orange!25}0.0400              \\ 
        
3D GS~\cite{kerbl20233d}  &               \cellcolor{orange!25}29.72      &        0.9129    &        0.1215        &                      23.59    &        0.9318        &         0.0631          &                       21.88       &     0.8847        &       0.1104            &    
       23.03      &    0.9583        &     0.0737   \\         
Ours  &              \cellcolor{red!25}40.43     &       \cellcolor{red!25}0.9812     &       \cellcolor{red!25}0.0267        &                      \cellcolor{red!25}42.14   &       \cellcolor{red!25}0.9937        &        \cellcolor{red!25}0.0063          &                      \cellcolor{red!25}36.43      &     \cellcolor{red!25}0.9838        &       \cellcolor{red!25}0.0174            &    
       \cellcolor{red!25}40.98      &    \cellcolor{red!25}0.9958     &    \cellcolor{red!25}0.0103            
        \\\midrule
                                 & \multicolumn{3}{c}{Lego} &                                    \multicolumn{3}{c}{T-Rex} &                                    \multicolumn{3}{c}{Stand Up} &                                   \multicolumn{3}{c}{Jumping Jacks}                                    \\
\cmidrule(lr){2-4} \cmidrule(lr){5-7} \cmidrule(lr){8-10} \cmidrule(lr){11-13}
Method                           & PSNR$\uparrow$           & SSIM$\uparrow$ & LPIPS$\downarrow$ & PSNR$\uparrow$            & SSIM$\uparrow$ & LPIPS$\downarrow$ & PSNR$\uparrow$               & SSIM$\uparrow$ & LPIPS$\downarrow$ & PSNR$\uparrow$                    & SSIM$\uparrow$ & LPIPS$\downarrow$ \\ \midrule
NeRF~\cite{mildenhall2021nerf}    & 20.30                    & 0.7900           & 0.2300              & 24.29                     & 0.9300           & 0.1300              & 18.19                        & 0.8900           & 0.1400              & 18.28                             & 0.8800           & 0.2300              \\
DirectVoxGo~\cite{sun2022direct}        & 21.13                    & 0.9000           & 0.1000              & 23.27                     & 0.9200           & 0.0900              & 17.58                        & 0.8600           & 0.1600              & 17.80                             & 0.8400           & 0.2000              \\
Plenoxels~\cite{fridovich2022plenoxels}& 21.97                    & 0.9000           & 0.1100              & 25.18                     & 0.9300           & 0.0800              & 18.76                        & 0.8700           & 0.1500              & 20.18                             & 0.8600           & 0.1900              \\
T-NeRF~\cite{pumarola2021d}       & 23.82                    & 0.9000           & 0.1500              & 30.19                     & 0.9600           & 0.1300              & 31.24                        & 0.9700           & 0.0200              & 32.01                             & 0.9700          & 0.0300              \\
D-NeRF~\cite{pumarola2021d}       & 21.64                    & 0.8300           & 0.1600              & \cellcolor{yellow!25}31.75                     & \cellcolor{yellow!25}0.9700           & \cellcolor{yellow!25}0.0300              & 32.79                        & 0.9800           & \cellcolor{yellow!25}0.0200              & \cellcolor{yellow!25}32.80                             & \cellcolor{yellow!25}0.9800           & \cellcolor{yellow!25}0.0300              \\
TiNeuVox-S~\cite{tineuvox}        & \cellcolor{yellow!25}24.35                    & 0.8800           & 0.1300              & 29.95                     & 0.9600           & 0.0600              & \cellcolor{yellow!25}32.89                        & \cellcolor{yellow!25}0.9800           & 0.0300              & 32.33                             & 0.9700           & 0.0400              \\
TiNeuVox-B~\cite{tineuvox}        & \cellcolor{orange!25}25.02                   & \cellcolor{yellow!25}0.9200           & \cellcolor{yellow!25}0.0700              & \cellcolor{orange!25}32.70                     & \cellcolor{orange!25}0.9800           & \cellcolor{orange!25}0.0300              & \cellcolor{orange!25}35.43                        & \cellcolor{orange!25}0.9900           & \cellcolor{orange!25}0.0200              & \cellcolor{orange!25}34.23                             & \cellcolor{orange!25}0.9800           & \cellcolor{orange!25}0.0300              \\ 

3D GS~\cite{kerbl20233d}  &               22.73      &        \cellcolor{orange!25}0.9282    &        \cellcolor{orange!25}0.0679       &                      21.92    &        0.9537        &         0.0498          &                       21.54       &     0.9283        &       0.0854            &    
       20.16      &    0.9279        &     0.0855   \\         
Ours  &              \cellcolor{red!25}25.16     &       \cellcolor{red!25}0.9451     &       \cellcolor{red!25}0.0421        &                      \cellcolor{red!25}37.25   &       \cellcolor{red!25}0.9923        &        \cellcolor{red!25}0.0115          &                      \cellcolor{red!25}43.35      &     \cellcolor{red!25}0.9929        &       \cellcolor{red!25}0.0092            &    
       \cellcolor{red!25}37.48      &    \cellcolor{red!25}0.9891      &    \cellcolor{red!25}0.0158       \\

\bottomrule
\end{tabular}%
}
\label{tab:synth2}
\end{table*}


\begin{table*}[h]
\centering
\caption{Per-scene quantitative results on real dynamic scenes. We color code each row as \colorbox{red!25}{best}, \colorbox{orange!25}{second best}, and \colorbox{yellow!25}{third best}.}
\resizebox{\textwidth}{!}{%
\begin{tabular}{lcccccccccccc}
\toprule
                                 & \multicolumn{3}{c}{torchocolate}                              & \multicolumn{3}{c}{cut-lemon}                                     & \multicolumn{3}{c}{chickchicken}                                          & \multicolumn{3}{c}{hand}                                     \\
\cmidrule(lr){2-4} \cmidrule(lr){5-7} \cmidrule(lr){8-10} \cmidrule(lr){11-13}
Method                           & PSNR$\uparrow$           & SSIM$\uparrow$ & LPIPS$\downarrow$ & PSNR$\uparrow$            & SSIM$\uparrow$ & LPIPS$\downarrow$ & PSNR$\uparrow$               & SSIM$\uparrow$ & LPIPS$\downarrow$ & PSNR$\uparrow$                    & SSIM$\uparrow$ & LPIPS$\downarrow$ \\ \midrule
NeRF~\cite{mildenhall2021nerf}    & 22.5                   & 0.866           & 0.373              & 24.1                    & 0.826           & 0.437              & 18.8                        & 0.761           & 0.453              & 23.8                             & 0.773           & 0.469              \\
NV~\cite{lombardi2019neural}        & 24.6                   & 0.917           & 0.189              & 28.8                     & \cellcolor{orange!25}0.951           & \cellcolor{red!25}0.190              & 22.6                        & 0.861           & 0.243              & 29.3                             & 0.912           & 0.213              \\
NSFF~\cite{li2021neural}& 22.3              & 0.883           & 0.253              & 28.0                     & 0.904           & 0.238              & 27.7                        & 0.939           & 0.173              & 24.9                             & 0.797           & 0.329              \\
Nerfies~\cite{park2021nerfies}       & \cellcolor{yellow!25}27.8                    & \cellcolor{orange!25}0.959           & \cellcolor{red!25}0.169              & \cellcolor{yellow!25}30.8                     & \cellcolor{yellow!25}0.946           & 0.223              & 28.7                        & \cellcolor{orange!25}0.948           & \cellcolor{red!25}0.141              & \cellcolor{yellow!25}29.9                             & \cellcolor{yellow!25}0.940           & \cellcolor{yellow!25}0.171              \\
HyperNeRF~\cite{park2021hypernerf}        & \cellcolor{orange!25}28.0                    & \cellcolor{red!25}0.962           & \cellcolor{orange!25}0.172              & \cellcolor{red!25}31.8                     & \cellcolor{red!25}0.956           & \cellcolor{yellow!25}0.210              & \cellcolor{yellow!25}28.7                        & \cellcolor{red!25}0.948           & \cellcolor{yellow!25}0.156              & \cellcolor{orange!25}30.7                             & \cellcolor{red!25}0.950           & \cellcolor{red!25}0.150              \\
TiNeuVox-S~\cite{tineuvox}        & 21.5                    & 0.754           & 0.478              & 23.4                   & 0.642           & 0.604              &     25.3                    & 0.761          & 0.485              &         24.2                    & 0.604           & 0.614             \\
TiNeuVox-B~\cite{tineuvox}        & 27.1                  & 0.824           & 0.395           & 28.6                 & 0.694          & 0.509             & \cellcolor{red!25}29.0                        & 0.812          & 0.408           & 27.3                          & 0.678          & 0.493             \\ 
        
3D GS~\cite{kerbl20233d}   &               21.8      &        0.787    &        0.402        &                      22.6    &        0.667        &         0.482          &                       20.2       &     0.719        &       0.517            &    
       23.6      &    0.723        &     0.351   \\         
Ours  &             \cellcolor{red!25}28.3	& \cellcolor{yellow!25}0.949 &	\cellcolor{yellow!25}0.174 &	\cellcolor{orange!25}31.4 &	0.945 &	\cellcolor{orange!25}0.205 &	\cellcolor{orange!25}28.8	& \cellcolor{yellow!25}0.942 &	\cellcolor{orange!25}0.146 &	\cellcolor{red!25}30.8 &	\cellcolor{orange!25}0.947 &	\cellcolor{orange!25}0.161 \\
\bottomrule
\end{tabular}%

}

\label{tab:real2}
\end{table*}

\end{document}